\documentclass{article}
\usepackage{graphicx,booktabs,amsmath,amssymb}
\usepackage{hyperref}
\usepackage[accepted]{icml2025}
\usepackage{iftex}
\ifXeTeX
  
  \DeclareFontShape{TU}{lmr}{m}{sl}{<->ssub*lmr/m/it}{}
  \DeclareFontShape{TU}{lmr}{bx}{n}{<->"[lmroman10-bold]:mapping=tex-text"}{}
\fi
\makeatletter
\def\Notice@String{Author preprint. November 2026.}
\makeatother
\hypersetup{pdftitle={Branch Caching After Step Distillation},pdfauthor={Mingyu Liang, Dingkun Xu},pdfsubject={Author preprint},colorlinks=true,linkcolor=black,citecolor=black,urlcolor=blue}
\icmltitlerunning{\raisebox{0pt}[6pt]{Branch Caching After Step Distillation}}
\begin{document}
\twocolumn[
\icmltitle{Branch Caching After Step Distillation\\ in Four- and Eight-Step Image Generators}
\begin{icmlauthorlist}
\icmlauthor{Mingyu Liang}{nju}
\icmlauthor{Dingkun Xu}{nju}
\end{icmlauthorlist}
\icmlaffiliation{nju}{School of Computer Science, Nanjing University, Nanjing, China}
\vskip 0.25in]
\printAffiliationsAndNotice{}\raggedbottom
\begin{abstract}
Can a distilled image generator save time by reusing features between its few remaining denoising steps? We study a fixed branch cache that replaces selected attention and MLP evaluations while preserving the student's noise schedule. One full trajectory and one raw-cache trajectory from the same initial noise supply a scalar correction for each branch type and reuse position. At $1024^2$, one internal reuse reduces four-step Qwen-Image-Lightning denoising time from 1.279 to 0.977 seconds. Three reuses yield $1.51\times$ speedup on eight-step Lightning and $1.53\times$ on eight-step Hyper-FLUX. We evaluate image quality on all 200 DrawBench prompts. With the reuse positions held fixed, correction improves mean same-seed image fidelity over raw reuse by 0.66 and 0.87 dB on the two Lightning students. On Hyper-FLUX the PSNR difference is 0.01 dB, while LPIPS improves from 0.1625 to 0.1546. These results measure the effect of branch caching and correction while retaining each student's original sampling grid.
\end{abstract}
\begin{table*}[t]
\centering
\caption{Additional speedup over the same uncached student at the same step count. Paired cells show uncached / calibrated means, with the better displayed value in bold. Latency is the mean denoising time measured on NVIDIA H800 GPUs. Speedups use paired H800 measurements. ImageReward and CLIP use all 200 DrawBench prompts.}
\label{tab:main}
\small\setlength{\tabcolsep}{4pt}\begin{tabular}{lccccc}\toprule
Student & Denoising (s) & Denoising speedup & Image-ready speedup & ImageReward & CLIP \\
\midrule
Lightning-4 & 1.279 / \textbf{0.977} & $1.309\times$ & $1.259\times$ & 1.2917 / \textbf{1.2980} & \textbf{32.5605} / 32.5591 \\
Lightning-8 & 2.527 / \textbf{1.671} & $1.513\times$ & $1.461\times$ & 1.2747 / \textbf{1.2922} & \textbf{32.7055} / 32.6572 \\
Hyper-FLUX-8 & 2.785 / \textbf{1.822} & $1.529\times$ & $1.475\times$ & 1.0250 / \textbf{1.0331} & \textbf{31.5979} / 31.5808 \\
\bottomrule\end{tabular}
\end{table*}

\section{Introduction}
Step distillation moves work from sampling into training. Once a four-step student has been released, an application can use its learned trajectory with only four transformer evaluations. Can the application also make one of those evaluations cheaper? Removing a scheduler update changes the trajectory on which the student operates. Retaining that update while approximating expensive internal branches leaves the student on its prescribed time grid.

This distinction matters at four steps. After two initial full evaluations, one can cache the third evaluation and still finish with a full evaluation. Even an ideal zero-cost cache would give only $4/3$ speedup under an equal-cost-step model. A practical implementation must pay for reading features, applying current gates, and writing residuals. And the reused features must survive a large change in noise level. For example, the first three Lightning-4 noise levels are $1.00$, $0.90$, and $0.75$. Agreement on adjacent steps in a long sampler gives little guidance about the third update here.

We examine a deployment question: given a released student, its original sampler, and one calibration prompt, how much latency and fidelity can we recover by reusing branch features? We adapt the branch correction developed in our CalibCache manuscript, \emph{Cache-to-Full Feature Dynamics Calibration for Diffusion Transformers}, to the student's own trajectory. The full-compute reference for calibration, quality comparisons, and timing is always the same distilled student. This choice isolates the additional effect of the cache.

We measure the deployment benefit against uncached inference. Corrected versus raw reuse, on identical steps, measures what calibration contributes.

\section{Related work}
\paragraph{Distilling a sampler.}
Progressive distillation trains a student to replace multiple teacher updates with one \citep{salimans}. Qwen-Image-Lightning provides step-specific distilled adapters for Qwen-Image \citep{lightning}. The Hyper-SD release includes a separate eight-step adapter for FLUX.1-dev \citep{hyper}. We take the released weights and guidance settings as fixed. No access to the original distillation data or training procedure is required.

\paragraph{Caching after distillation.}
PostDiff studies mixed-resolution inference and module caching on models including a latent consistency model \citep{postdiff}. DisCa trains a feature predictor for step-distilled video transformers \citep{disca}. We study the incremental effect of this correction on two released image-model families.

\paragraph{Predicting intermediate features.}
DeepCache reuses high-level U-Net features \citep{deepcache}. FORA periodically reuses attention and MLP outputs in diffusion transformers \citep{fora}. This provides the raw-reuse control for our design. TaylorSeer forecasts features from their history \citep{taylor}. With only two full observations before our first reuse, we estimate one scalar per branch type and position from the difference between the full and raw-reuse trajectories.

\begin{figure*}[t]
\centering
\includegraphics[width=.80\textwidth]{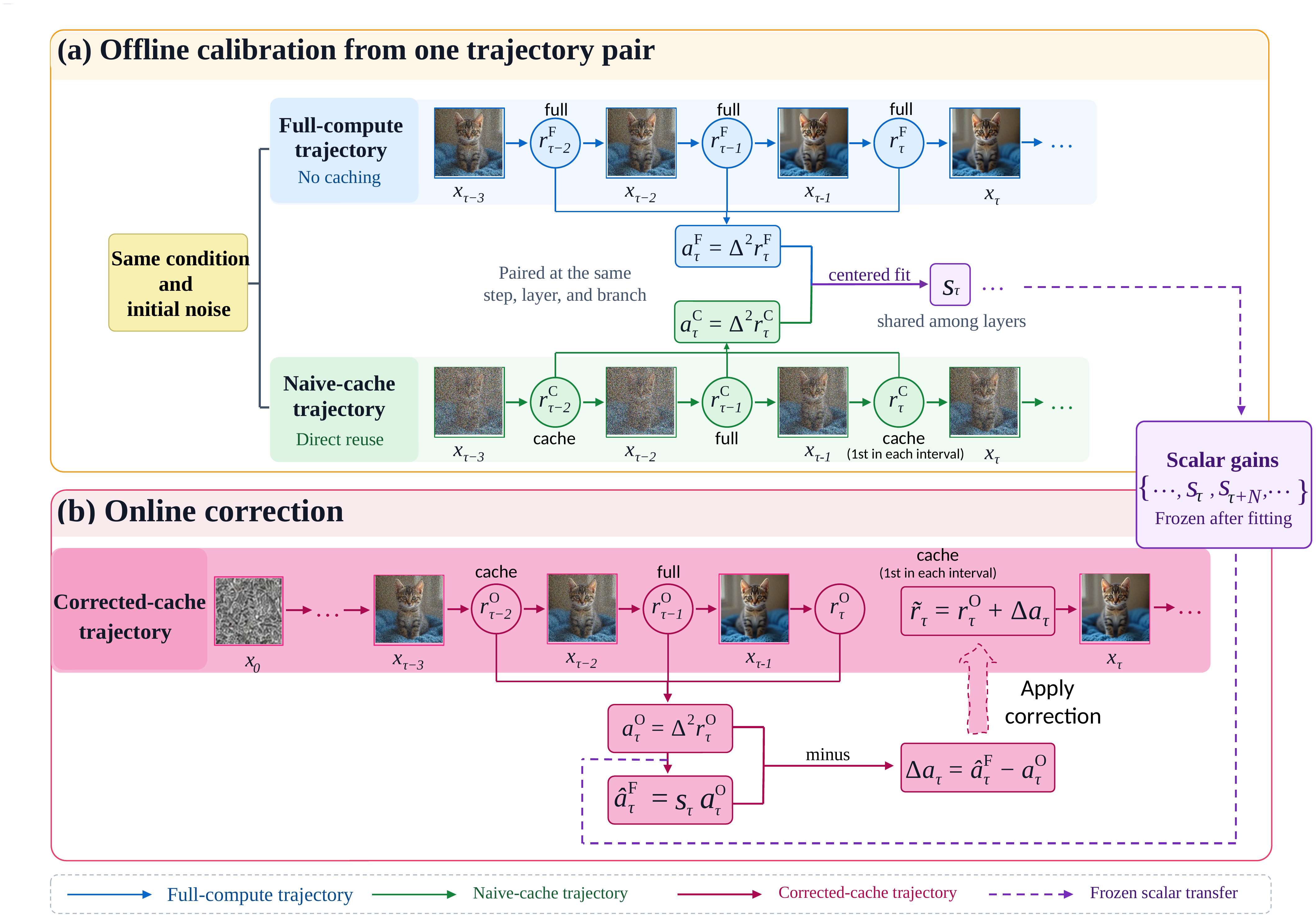}
\caption{Branch calibration and reuse. Reproduced from our CalibCache manuscript. The top panel pairs full and raw-cache trajectories from the same initial noise. The lower panel applies the fitted gain to the current branch's second difference. The diagram illustrates a reuse after an earlier cache step. In the four-step setting the correction runs once, at index 2, after two full evaluations. The eight-step schedule applies it at indices 2, 4 and 6.}
\label{fig:calibcache-method}
\end{figure*}

\section{A branch cache for short schedules}
Figure~\ref{fig:calibcache-method} shows how one calibration pair supplies the branch coefficients used throughout generation.
\paragraph{Where computation is reused.}
Consider Lightning-4. Steps 0 and 1 compute all branches. At step 2, each cached branch starts with its output from step 1 and adds a fitted multiple of its change from step 0 to step 1. The current gates, residual additions and output head turn those features into a new solver input. Step 3 runs in full on the resulting latent. Thus a call saved in the middle of a four-step trajectory can affect the last full evaluation. At eight steps we reuse at indices $2,4,6$ and compute $0,1,3,5,7$ in full. All indices execute a scheduler update.

For Lightning we cache four pre-gate outputs: image and text attention, and image and text MLP. Hyper-FLUX uses the same four types in double-stream blocks. In single-stream blocks the two branch outputs share a final projection. We record the projected attention contribution and recover the MLP contribution from the joint projection after accounting for the bias. At reuse, the two branch corrections are added to the cached joint projection. This preserves the original projection and its bias on the full path.

\paragraph{Calibrating the amount of extrapolation.}
Let $r_t$ denote the raw branch output used at index $t$ in one block. At a cached position $r_t=r_{t-1}$ because the preceding position is full. Form the step-index second difference
\begin{equation}
 a_t=r_t-2r_{t-1}+r_{t-2}.
\end{equation}
The method uses index differences even on a nonuniform noise grid. For branch type $b$, return
\begin{equation}
 \widehat r_t=r_t+(s_{t,b}-1)a_t. \label{eq:corr}
\end{equation}
The residual path receives $\widehat r_t$ and the history buffer stores $r_t$. Consequently, in real arithmetic Eq.~\ref{eq:corr} equals $r_{t-1}+(1-s_{t,b})(r_{t-1}-r_{t-2})$. A coefficient of one recovers raw reuse. The first Lightning-8 cache position, for example, has fitted coefficients of 0.908 for image attention and 0.221 for the image MLP. The latter extrapolates farther along its observed change. Later cache positions use the intervening raw history, which can itself contain reused values.

Use the same calibration prompt and initial noise for two rollouts. One computes every branch and supplies a full-trajectory second difference $Y$. The other uses raw reuse and supplies its own second difference $X$. The two latent trajectories evolve independently after the first approximation. Pool token rows across layers of the same branch type at a fixed step and subtract each channel's sample mean. With centered arrays $X_c,Y_c$, fit
\begin{equation}
 s_{t,b}=\frac{\langle X_c,Y_c\rangle_F}
 {\max(\|X_c\|_F^2,10^{-12})}.
\end{equation}
This is the least-squares slope for the centered statistics when the floor is inactive. Inference uses the slope with zero bias. All generated prompts share the same coefficients for that checkpoint and schedule. There are four fitted scalars for Lightning-4, twelve for Lightning-8 and eighteen for Hyper-FLUX-8 for the schedules used here.

\section{Experiments}
\paragraph{Inputs and controls.}
For image quality evaluation, we use all 200 DrawBench prompts \citep{saharia}, one CUDA seed $314159+i$ for prompt index $i$, BF16, $1024\times1024$ resolution and batch size one. Lightning-4 uses the V1.0 adapter and Lightning-8 the V1.1 adapter, fused at scale one with external CFG one and the shift-three Euler schedule. Hyper-FLUX-8 uses FLUX.1-dev, the eight-step Hyper-SD LoRA fused at 0.125, guidance 3.5 and its released Euler configuration. Each configuration is calibrated separately on ``A red ceramic cup on a wooden table near a window'' with seed 364159. Calibration has no access to DrawBench scores. A native run and a wrapped all-full run were checked for latent equality on a calibration-independent test input.

\paragraph{What the measurements capture.}
ImageReward \citep{ir} and CLIP ViT-B/32 cosine similarity scaled by 100 \citep{clip} score prompt--image pairs. For fidelity we compare each output to its same-seed uncached student. We average full-resolution RGB SSIM \citep{ssim} and AlexNet LPIPS v0.1 \citep{lpips} over images. SSIM uses a valid $11\times11$ Gaussian window ($\sigma=1.5$), population moments and unit-range RGB. LPIPS takes float32 RGB in $[-1,1]$. Higher SSIM and lower LPIPS indicate closer agreement with the reference.

Denoising time spans the first transformer call through the final sampling update. Image-ready time spans the resident pipeline call through decoded RGB. We report mean denoising times measured on NVIDIA H800 GPUs. We warm the inference paths and rotate their timing order. Each prompt's paired runs use the same GPU. The timed call excludes loading, adapter fusion, offline fitting, first-use compilation and file writes. Optimized outputs match the archived corrected RGB arrays.

\paragraph{How much time remains to save?}
Table~\ref{tab:main} measures the remaining deployment benefit after distillation. Denoising savings are 0.302, 0.857, 0.963 s in table order. The previous implementations took 1.008, 1.928, 2.091 s at the same refresh positions. The optimized implementation improves on those times by $1.03\times$, $1.15\times$, $1.15\times$. Every timed output matches its archived counterpart.

Lightning can reuse timestep modulation even when computing every branch. Branch caching achieves $1.2773\times$ and $1.4740\times$ speedup over the all-full modulation control at four and eight steps. The paired 95\% prompt-bootstrap intervals are $[1.2771,1.2776]$ and $[1.4733,1.4746]$.

\paragraph{How tightly are quality changes measured?}
For Lightning-4, Lightning-8 and Hyper-FLUX-8 respectively, the paired 95\% intervals for the change in ImageReward are $[-.004,.017]$, $[-.004,.045]$, and $[-.020,.037]$. For CLIP they are $[-.086,.084]$, $[-.143,.054]$, and $[-.145,.117]$. Each interval comes from 10,000 paired prompt bootstrap resamples (seed 20261115) and is unadjusted for multiple comparisons. The data support small observed mean changes on this prompt set. Statistical equivalence would require a specified margin and a corresponding test.

\begin{table*}[t]
\centering\small
\caption{Cache comparisons at the original scheduler steps. Quality metrics use DrawBench-200. Latency is the mean denoising time measured on NVIDIA H800 GPUs. Speedup is the ratio of mean times for each method and its own paired native run. Full counts full transformer evaluations per image. PSNR pools RGB squared errors against each same-seed native image before conversion to dB. Bold marks the best reported accelerated values at displayed precision. IR denotes ImageReward.}
\label{tab:external}
\setlength{\tabcolsep}{3.5pt}\begin{tabular}{llrrrrrrrr}\toprule
Student & Method & Full & Sec.$\downarrow$ & Speedup$\uparrow$ & IR$\uparrow$ & CLIP$\uparrow$ & PSNR$\uparrow$ & SSIM$\uparrow$ & LPIPS$\downarrow$ \\
\midrule
Lightning-4 & Native & 4 & 1.284 & 1.00 & 1.292 & 32.56 & $\infty$ & 1.0000 & 0.0000 \\
  & TeaCache & 3.04 & 1.030 & 1.28 & 1.285 & \textbf{32.66} & 21.99 & 0.8290 & 0.0969 \\
  & TaylorSeer & 3.00 & 1.045 & 1.22 & 1.297 & 32.51 & 21.63 & 0.8418 & 0.0874 \\
  & HiCache & 3.00 & 1.034 & 1.21 & 1.296 & 32.52 & 21.74 & 0.8424 & 0.0865 \\
  & CalibCache & 3.00 & \textbf{0.980} & \textbf{1.31} & \textbf{1.298} & 32.56 & \textbf{22.27} & \textbf{0.8505} & \textbf{0.0783} \\
\midrule
Lightning-8 & Native & 8 & 2.494 & 1.00 & 1.275 & 32.71 & $\infty$ & 1.0000 & 0.0000 \\
  & TeaCache & 5.21 & 1.877 & 1.42 & 1.279 & \textbf{32.69} & 19.17 & 0.7954 & 0.1358 \\
  & TaylorSeer & 5.00 & 1.875 & 1.36 & 1.280 & 32.65 & 18.95 & 0.8081 & 0.1238 \\
  & HiCache & 5.00 & 1.876 & 1.33 & 1.271 & 32.62 & 18.92 & 0.7999 & 0.1270 \\
  & CalibCache & 5.00 & \textbf{1.650} & \textbf{1.51} & \textbf{1.292} & 32.66 & \textbf{19.51} & \textbf{0.8228} & \textbf{0.1121} \\
\midrule
Hyper-FLUX-8 & Native & 8 & 2.776 & 1.00 & 1.025 & 31.60 & $\infty$ & 1.0000 & 0.0000 \\
  & TeaCache & 5.13 & 1.950 & 1.49 & 1.019 & \textbf{31.65} & 20.85 & 0.7757 & 0.1912 \\
  & TaylorSeer & 5.00 & 1.905 & 1.47 & 1.029 & 31.64 & 20.43 & 0.7798 & 0.1752 \\
  & HiCache & 5.00 & 1.932 & 1.43 & 1.021 & 31.57 & 20.63 & 0.7877 & 0.1659 \\
  & CalibCache & 5.00 & \textbf{1.817} & \textbf{1.53} & \textbf{1.033} & 31.58 & \textbf{21.24} & \textbf{0.7946} & \textbf{0.1546} \\
\bottomrule\end{tabular}
\end{table*}

\paragraph{External cache baselines.}
We adapt three published methods to the frozen students. TaylorSeer~\citep{taylor} and HiCache~\citep{hicache} use order-two branch forecasts, with the available order reduced automatically at early anchors. HiCache uses its scaled Hermite basis at scale 0.6. Both retain the two initial full steps and interval two of our main setting. On FLUX single-stream blocks they forecast the joint projection; on double-stream blocks they forecast the separate branches.

TeaCache~\citep{tea} monitors the relative change in the first block's modulated image input and reuses the transformer-stack residual. We transfer the published FLUX polynomial and the vLLM-Omni Qwen polynomial. Four calibration prompts with seeds 364159--364162 select the threshold whose mean skip count is closest to our budget. The first and last evaluations remain full. Actual test-time counts can differ from the calibration target.

The fitted TeaCache thresholds are 6.1875, 0.753125 and 0.55625 in table order. CalibCache uses one calibration pair. The amount of work on a cached update varies by method, so equal full-call counts can yield different wall times.

\paragraph{Speed and fidelity across cache methods.}
On Lightning-8, the mean ImageReward of CalibCache is 1.292, compared with 1.279 for TeaCache, 1.280 for TaylorSeer and 1.271 for HiCache. The 95\% paired interval for its difference from TaylorSeer is $[-.001,.025]$. On Hyper-FLUX-8, CalibCache attains LPIPS 0.1546 and HiCache attains 0.1659. TeaCache has the highest mean CLIP score on that student.

\section{What does calibration add?}
\begin{table}[t]
\centering\small
\caption{Paired fidelity relative to each uncached student. Both methods reuse the same branches at the same steps. Each image contributes one dB value, whereas Table~\ref{tab:external} pools squared errors. Intervals are paired prompt-bootstrap intervals for corrected minus raw.}
\label{tab:fid}
\setlength{\tabcolsep}{2.5pt}\begin{tabular}{lrrl}\toprule
Student & Raw & Corrected & Difference [95\% CI] \\
\midrule
Lightning-4 & 23.30 & \textbf{23.97} & $+.66\ [+.59,+.74]$ \\
Lightning-8 & 20.68 & \textbf{21.55} & $+.87\ [+.74,+1.01]$ \\
Hyper-FLUX-8 & 22.62 & \textbf{22.63} & $+.01\ [-.04,+.07]$ \\
\bottomrule\end{tabular}
\par\smallskip\begin{tabular}{lcc}\toprule
Student & SSIM (raw / corr.) & LPIPS (raw / corr.) \\
\midrule
Lightning-4 & 0.8273 / \textbf{0.8505} & 0.0960 / \textbf{0.0783} \\
Lightning-8 & 0.7882 / \textbf{0.8228} & 0.1356 / \textbf{0.1121} \\
Hyper-FLUX-8 & 0.7923 / \textbf{0.7946} & 0.1625 / \textbf{0.1546} \\
\bottomrule\end{tabular}
\end{table}
The raw control keeps the step schedule, student and gate computation fixed and sets $s=1$. We decoded the three archived RGB outputs per prompt, verified their generation hashes, and computed PSNR separately for raw and calibrated reuse against the native image. Table~\ref{tab:fid} averages per-image dB. The correction improves PSNR on 192/200 Lightning-4 prompts and 179/200 Lightning-8 prompts. On Hyper-FLUX-8 it improves 91/200, with a mean difference close to zero. These intervals use 10,000 paired prompt resamples (seed 20261118). They condition on the fitted coefficients and generation seeds.

\begin{figure}[t]
\centering\setlength{\tabcolsep}{1pt}\scriptsize
\begin{tabular}{@{}lccc@{}}
 &Native&Raw reuse&Calibrated\\
L4&\includegraphics[width=.265\columnwidth]{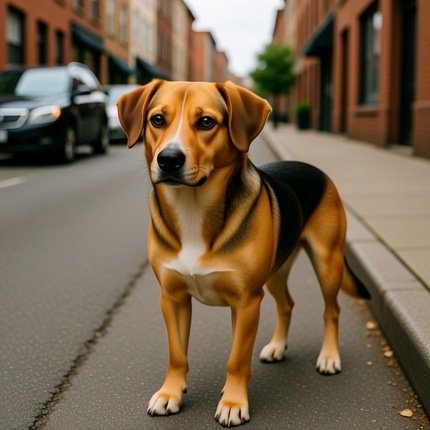}&\includegraphics[width=.265\columnwidth]{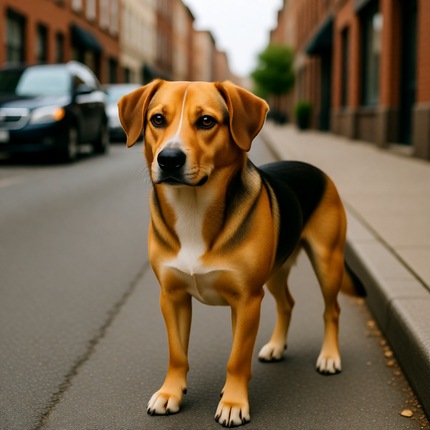}&\includegraphics[width=.265\columnwidth]{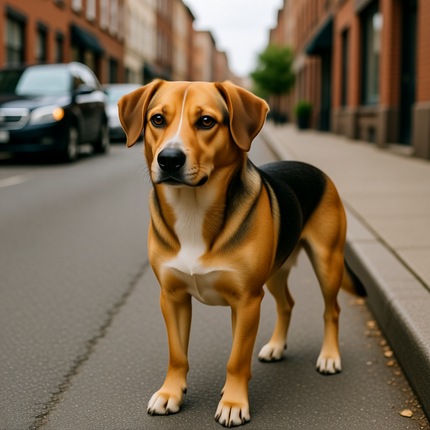}\\
L8&\includegraphics[width=.265\columnwidth]{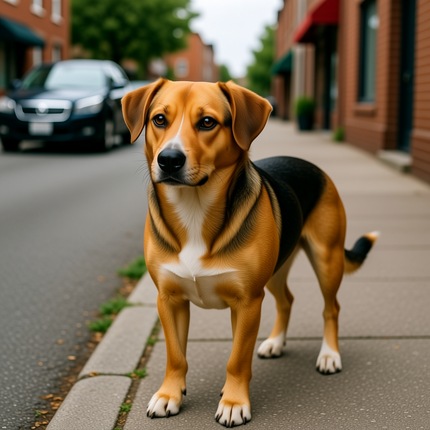}&\includegraphics[width=.265\columnwidth]{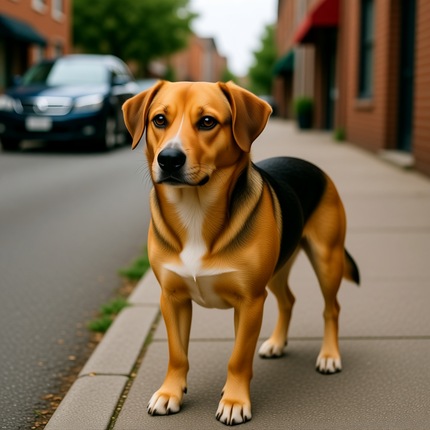}&\includegraphics[width=.265\columnwidth]{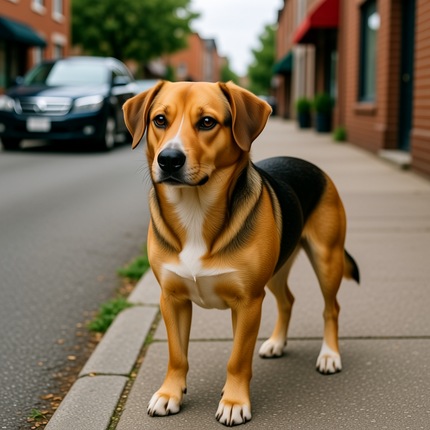}\\
H8&\includegraphics[width=.265\columnwidth]{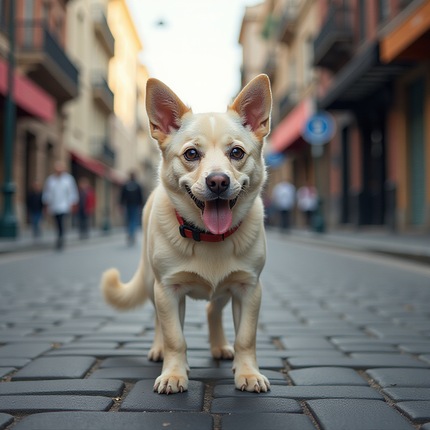}&\includegraphics[width=.265\columnwidth]{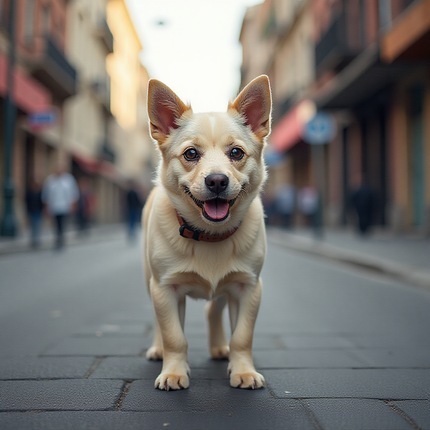}&\includegraphics[width=.265\columnwidth]{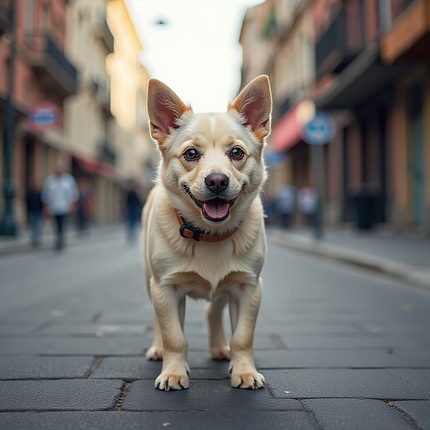}
\end{tabular}
\caption{``One dog on the street.'' DrawBench index 40, seed 314199, from the fixed preview indices. Rows are Lightning-4, Lightning-8 and Hyper-FLUX-8. The broad composition survives in each row. In H8, raw reuse smooths the foreground paving visible in the native image. Images are downsampled for display.}
\label{fig:dog}
\end{figure}

Calibration increases mean SSIM over raw reuse by 0.0232, 0.0346 and 0.0023, and reduces LPIPS by 0.0177, 0.0234 and 0.0079 in table order. Thus Hyper-FLUX has a small structural and perceptual improvement despite its near-zero mean PSNR change. Prompt-based metrics give a different view. Relative to raw reuse, calibration changes mean ImageReward by $+.0104$, $+.0129$ and $+.0193$ in table order. CLIP changes by $-.0983$, $-.0180$ and $-.0233$. For Lightning-4, even the paired interval for the CLIP change lies below zero ($[-.151,-.045]$). A closer reproduction of the student's chosen image can therefore score lower on a text--image metric. Figure~\ref{fig:dog} shows the three outputs on a common prompt.

\section{Discussion}
Four-step Lightning saves 0.302 s per denoising run. At the same refresh positions, correction improves PSNR, SSIM and LPIPS over raw reuse on both Lightning students. On Hyper-FLUX, the mean PSNR change is near zero and LPIPS decreases.

The intervals resample prompts at fixed generation seeds. Lightning uses different adapter versions at four and eight steps, which confounds a comparison of step counts. The timing results cover resident inference. Cold-start payback, concurrent throughput and memory-constrained use remain unmeasured. Calibration-noise sensitivity and independently latency-matched operating points remain open.

{\small\setlength{\bibsep}{2pt}
}
\appendix\small
\section{A 50-prompt DetailMaster Mini evaluation}
\label{app:detail}
We applied the existing two-initial-step Lightning packs to a frozen subset of DetailMaster Mini \citep{detail}. The release contains 800 designated-task rows. The 50 selected descriptions span 178--396 words (median 254.5). We sampled six or seven prompts per task using deterministic category shuffles, deduplicated source image identities, and fixed the inputs before inspecting generated images. At each step count, both methods used the same 50 prompts and seeds $314159$--$314208$. This yields 200 images across both checkpoints. All main-text timings come from DrawBench.

\begin{table}[h]
\centering\small
\caption{Designated-task accuracy (\%) on the Mini subset. Columns give native / calibrated Lightning. Each prompt is scored only for its designated task. Macro averages give equal weight to the eight task percentages.}
\begin{tabular}{lcc}\toprule
Task & Four steps & Eight steps\\\midrule
Object &\textbf{66.25} / 61.25&\textbf{65.00} / 56.25\\
Animal &28.07 / \textbf{45.61}&\textbf{24.56} / 21.05\\
Person &23.29 / \textbf{28.77}&27.40 / \textbf{30.14}\\
Location &48.00 / \textbf{52.00}&\textbf{52.00} / 48.00\\
Background &\textbf{85.71} / \textbf{85.71}&85.71 / \textbf{100.00}\\
Lighting &\textbf{100.00} / \textbf{100.00}&\textbf{100.00} / \textbf{100.00}\\
Style &\textbf{83.33} / \textbf{83.33}&\textbf{83.33} / \textbf{83.33}\\
Spatial &48.65 / \textbf{51.35}&48.65 / \textbf{51.35}\\\midrule
Macro &60.41 / \textbf{63.50}&60.83 / \textbf{61.27}\\\bottomrule
\end{tabular}
\end{table}

The release evaluator combines object localization and vision--language scoring. The recorded code revision is 4f53b955; we changed bounding-box parsing to safe literal evaluation. An exception in the release's BLIP fallback can count unmatched detections as absent. In one eight-step image pair the confidence for a lavender circle changed from 0.262 to 0.208 across methods. The default threshold of 0.25 discarded the latter detection and all eight associated attributes. As a post hoc sensitivity check, lowering the confidence threshold to 0.20 and the box agreement threshold from 0.70 to 0.65 changed object accuracy to 66.25/68.75 at four steps and 65.00/66.25 at eight steps. The default results are reported in the table. The direction of this task-level comparison depends on evaluator settings.

\begin{table}[ht]
\centering\small
\caption{DetailMaster Mini subset. Equal-weight mean designated-task accuracy (\%) on the identical 50 prompts using the default evaluator. Acc. averages the eight task scores equally. SSIM and LPIPS average over images against the matching native output. Bold compares accelerated methods.}
\label{tab:external-mini}
\scriptsize\setlength{\tabcolsep}{2.1pt}\begin{tabular}{lrrr rrr}\toprule
 & \multicolumn{3}{c}{Lightning-4} & \multicolumn{3}{c}{Lightning-8} \\
Method & Acc. & SSIM & LPIPS & Acc. & SSIM & LPIPS \\
\midrule
Native & 60.41 & 1.0000 & 0.0000 & 60.83 & 1.0000 & 0.0000 \\
TeaCache & 62.05 & 0.8455 & 0.1093 & \textbf{62.66} & 0.8178 & 0.1584 \\
TaylorSeer & 62.50 & 0.8555 & 0.1030 & 61.82 & 0.8221 & 0.1508 \\
HiCache & \textbf{63.52} & 0.8568 & 0.1011 & 61.21 & 0.8167 & 0.1537 \\
CalibCache & 63.50 & \textbf{0.8731} & \textbf{0.0921} & 61.27 & \textbf{0.8363} & \textbf{0.1379} \\
\bottomrule\end{tabular}
\end{table}

Table~\ref{tab:external-mini} applies the unchanged default Mini evaluator to all three external methods. CalibCache and HiCache differ by 0.02 percentage points at four steps. At eight steps TeaCache has the highest observed macro average (62.66\%); CalibCache scores 61.27\%. The supplementary results include per-task numerators and denominators. A complementary blinded paired evaluation used Gemini 3.8 Flash, one judgment per pair, with calibrated images on the left for 25 of 50 pairs at each step count. Mean relative scores (native / calibrated, scaled by 100) were 50.12/49.88 for prompt alignment, 50.52/49.48 for visual quality and 50.40/49.60 for overall preference at four steps. The eight-step values were 49.96/50.04, 50.48/49.52 and 50.16/49.84. All 100 responses were valid. These are automated relative scores on a small subset, with no human preference estimate.

\begin{figure}[h]
\centering\small
\begin{tabular}{@{}cc@{}}
Native & Calibrated\\
\includegraphics[width=.46\columnwidth]{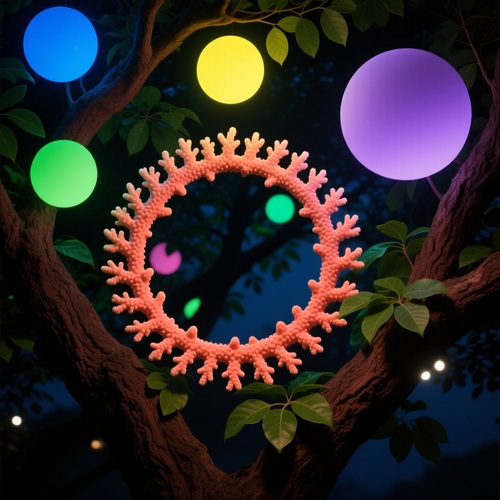} &
\includegraphics[width=.46\columnwidth]{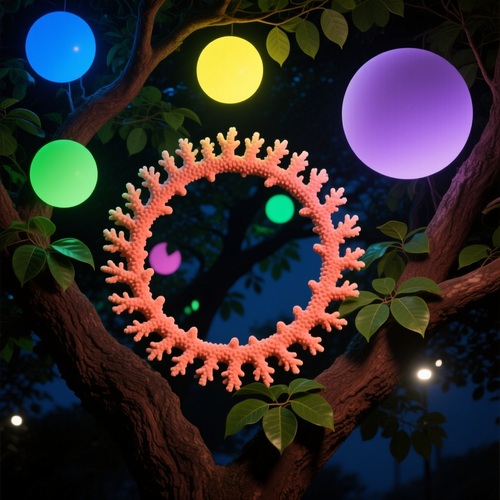}
\end{tabular}
\caption{DetailMaster Mini subset image 032, eight-step Lightning. The prompt describes colored lights in a tree with specified positions and attributes. This example was selected to illustrate the detector-threshold case in the text.}
\end{figure}
\section{Reproduction details}
The Lightning adapters (V1.0 at four steps, V1.1 at eight) use revision d468c53. Hyper-FLUX uses Hyper-SD revision bc08d97, FLUX.1-dev and maximum text length 512.

Lightning fuses the cached block sequence and reuses modulation only for identical embeddings. It casts the correction to BF16 before addition. Hyper-FLUX stores two BF16 branch observations, corrects in FP32 and fuses current gates with residual updates. Its final full step omits history writes.
\end{document}